\pdfoutput=1
\documentclass[11pt]{article}

\usepackage[final]{acl}

\usepackage{times}
\usepackage{latexsym}
\usepackage[T1]{fontenc}
\usepackage[utf8]{inputenc}
\usepackage{microtype}

\usepackage{graphicx}
\usepackage{booktabs}       %
\usepackage{multirow}
\usepackage{amsmath}
\usepackage{xcolor}
\usepackage{enumitem}       %
\usepackage{listings}       %
\usepackage{makecell}       %

\graphicspath{{figures/}}
\usepackage[most]{tcolorbox} %
\usepackage{tabularx}
\newtcblisting{promptbox}{listing only, breakable, colback=gray!6, colframe=gray!45,
  boxrule=0.4pt, arc=1pt, left=4pt, right=4pt, top=3pt, bottom=3pt,
  listing options={basicstyle=\footnotesize\ttfamily, columns=fullflexible, keepspaces=true,
    breaklines=true, breakatwhitespace=true, breakindent=0pt, upquote=true}}

\newcommand{\lingex}[1]{\textit{#1}} %

\newif\ifnotes \notestrue

\title{Evaluating Large Language Models on conversational African American English: Feature classification and next-turn generation}

\author{
  Romeo De Timmerman \\
  Columbia University \\
  Department of Computer Science
  \And
  Thomas Demeester \\
  Ghent University -- imec \\
  IDLab
  \And
  Els Lefever \\
  Ghent University \\
  LT3
}

\begin{document}
\maketitle

\begin{abstract}
Large Language Models (LLMs) are known to misinterpret and misrepresent regional and social varieties of English other than Mainstream American English (MAE), such as African American English (AAE), yet most evaluations of these biases have relied on synthetic, stylized or decontextualized data.
This paper instead evaluates LLMs on authentic, conversational AAE to more closely reflect how such models may encounter AAE when deployed as chatbots.
Concretely, we perform two complementary evaluation tasks on 100 six-turn segments randomly sampled from the Corpus of Regional African American Language \citep[CORAAL;][]{KendallFarrington2024}, comparing three closed-weight LLM configurations: Gemini 2.5 Flash with and without reasoning, and GPT-4.1 mini.
The first task compares the models against gold-standard manual annotations for five AAE morphosyntactic features, while the second assesses their ability to generate contextually appropriate next turns, evaluating outputs for AAE feature frequency, diversity (Self-BLEU), and pragmatic quality.
Reasoning-enabled Gemini is the strongest feature detector by a clear margin (macro F1 of 0.91), outperforming both other LLMs as well as a finetuned BERT-based baseline \citep{MasisEtAl2022}.
In generation, however, all LLMs underproduce AAE relative to the human speakers and produce relatively uniform continuations, with reasoning-enabled Gemini producing the least AAE and the least diverse output.
We conclude that a model's ability to detect AAE does not straightforwardly translate into its spontaneous use of AAE in generation, and argue that addressing this gap requires changes upstream, in how pretraining and alignment data represent dialectal variation.
\end{abstract}

\section{Introduction}
\label{sec:introduction}

This paper builds on a growing body of scholarly work in Natural Language Processing (NLP) that has laid bare systematic biases in how language models interpret and handle underrepresented linguistic varieties.
These concerns are especially salient for African American English (AAE)\footnote{The variety is variously labeled African American English (AAE), African American Vernacular English (AAVE) or, increasingly, African American Language (AAL) \citep[cf.][]{King2020}. We use AAE throughout, including where cited work adopts a different label.}, a variety of English spoken predominantly, though not exclusively, by African Americans across the United States.
Despite decades of sociolinguistic research documenting its grammatical and pragmatic richness, AAE remains underrepresented in Large Language Model (LLM) training data \citep{DeasEtAl2025} and resources for model finetuning \citep{MireEtAl2025}.
When such models are deployed in downstream applications such as automated customer-service systems, content moderation pipelines or educational technologies, misinterpretation of AAE can restrict access to services, impede communication or unfairly penalize users.

Where previous research has mostly evaluated LLM performance disparities against AAE using synthetic or contextually constrained data (cf.\ Section~\ref{sec:background}), this paper introduces a conversational approach, drawing on the Corpus of Regional African American Language \citep[CORAAL;][]{KendallFarrington2024}, a collection of sociolinguistic interviews with AAE speakers which captures spontaneous interactional speech across multiple regions and generations.
From this corpus, we sample conversational excerpts and use them as input for two complementary LLM evaluation tasks.
The first task addresses whether LLMs can correctly detect AAE morphosyntactic features in conversational excerpts, evaluated against human annotation and a supervised classifier.
The second task addresses how LLMs continue a conversational excerpt, and whether they accommodate to an AAE-speaking interlocutor or default to Mainstream American English (MAE) regardless of context.

\section{Background}
\label{sec:background}

AAE is a systematic, rule-governed and regionally diverse variety of English with phonological, morphological and syntactic features that differ from those of MAE, such as copula deletion (e.g.\ \lingex{she fun}), which involves the deletion of a form of the verb \lingex{to be}, and habitual be (e.g.\ \lingex{he be walking to school}), in which an invariant form of the verb \lingex{to be} encodes a habitual action.
AAE's status as a rule-governed system rather than a `deficient' form of MAE has been established since \citeposs{Labov1972} seminal work on the variety.
Subsequent work has stressed that AAE is not monolithic, but, like MAE, varies across contexts, regions and generations of speakers \citep{King2020}.

Despite the rich body of scholarly work on AAE, its status as a fully-fledged linguistic system remains contested outside of linguistics.
Scholars have consequently documented material consequences for speakers of AAE in domains such as education, where AAE-speaking children are routinely assessed against MAE norms (\citealp{CharityEtAl2004}; see also the Ann Arbor Black English trial and the Oakland Ebonics controversy), and the courtroom, where reporters have been shown to transcribe AAE speech far less accurately than their certification requires \citep{JonesEtAl2019,RickfordKing2016}.

More recently, a similar pattern has surfaced in NLP, where language models often display measurable performance disparities when confronted with AAE input.
Crucially, as LLMs shape more and more aspects of public life, the consequences of misinterpreting and mishandling dialectal features extend beyond computational accuracy in an abstract sense.
On the one hand, misclassification and misrepresentation of AAE constitute representational harms \citep{BlodgettEtAl2020}, as models encode language ideologies that stigmatize minoritized speech varieties.
Such patterns reinforce longstanding hierarchies by portraying AAE as deviant, undesirable or otherwise different from the default norm.
On the other hand, these representational failures can in turn produce allocational harms \citep{BlodgettEtAl2020}, resulting in direct material consequences when LLMs mediate access to information, services and opportunities.

A growing body of work in NLP and computational linguistics has consequently begun to evaluate language model performance on AAE input, showing where biases may occur, while also offering initial avenues for improved model development.
Much of this early work has focused on classification tasks, where off-the-shelf tools have been shown to consistently underperform on AAE data in language identification, part-of-speech tagging, dependency parsing and hate speech detection \citep{BlodgettOConnor2017,BlodgettEtAl2018,JorgensenEtAl2016,HarrisEtAl2022}.
Additionally, AAE features which do not occur in MAE, such as habitual be, have been shown to impair both speech recognition \citep{MartinTang2020} and automatic feature classification \citep{PrevilonEtAl2024}.

Recent scholarship has mostly focused on how LLMs handle AAE during generation.
\citet{DeasEtAl2023}, for instance, report a counterpart generation task, in which models produce an AAE text given an MAE one and vice versa, showing that models routinely fail to preserve the meaning and tone of AAE input.
Similarly, \citet{WuEtAl2026} demonstrate that instruction-tuned LLMs systematically prefer MAE continuations even when the preceding context is in AAE, with syntactic patterns such as negative concord acting as the strongest triggers.
A parallel strand of work has proposed evaluating the overall pragmatic quality of AAE generated by LLMs.
\citet{FleisigEtAl2024} report that explicitly instructing a model to match a speaker's variety increases rather than reduces stereotyping, which suggests that simply prompting models to accommodate dialectal variation is not a sufficient solution.
Moreover, \citet{BasoahEtAl2025} show that African American users of AI writing tools experience them as alienating, both when they fail to recognize AAE, as well as when they attempt to imitate it.
Concerns for LLMs producing features of AAE are also reported in \citet{FinchEtAl2025}, who find that AAE-speaking evaluators rate text-based chatbots less favorably as the density of AAE features in their output rises.

To measure and document these performance disparities, several scholars have constructed aligned corpora: collections of texts where each entry has both an AAE and an MAE version.
Most of these efforts have focused on creating aligned datasets by generating synthetic AAE text from authentic MAE using rule-based transformations \citep{ZiemsEtAl2022,DeasEtAl2024}, or by prompting LLMs to rewrite MAE input into AAE \citep{GravesEtAl2024}.
Such approaches provide valuable datasets for evaluating bias, but they remain limited by their reliance on synthetic AAE, which may reproduce stereotyped or inauthentic linguistic patterns that are not representative of human-produced AAE.
This is supported by \citet{LinEtAl2025}, who report that injecting rule-based AAE features into reasoning benchmarks degrades model performance considerably less than human-written AAE rewrites of the same items.
In other words, synthetic rewrites of MAE into AAE are useful, but not unproblematic proxies.

A small number of studies have instead worked with authentic AAE source material, either paired with human-translated MAE equivalents \citep{Dacon2022,DeasEtAl2023,GroenwoldEtAl2020} or used directly as model input without an MAE counterpart \citep{FleisigEtAl2024}.
Such material, however, has mostly consisted of social media posts or hip hop lyrics, which are often highly stylized or performative.
\citet{DeasEtAl2023} is a notable exception, combining social media and lyrics with interview and focus group data, though the latter are sampled as single utterances stripped of their surrounding context.
While these studies have identified systematic biases against AAE in LLMs, the synthetic, stylized or contextually constrained data they relied on may still not accurately reflect how speakers use AAE in everyday conversation, or, more importantly, how models encounter it when responding to user input.
Moreover, \citet{DeasEtAl2025} show that even authentic AAE text found in pretraining corpora is often performative and underrepresents ordinary spoken dialogue.
This paper therefore turns to spontaneous, conversational AAE, which more closely resembles the dyadic, turn-by-turn exchanges through which LLMs, especially deployed as chatbots, may encounter AAE.

\section{Data and models}
\label{sec:data-and-models}

\subsection{CORAAL}
\label{sec:coraal}

This paper draws on dialogue data extracted from The Corpus of Regional African American Language \citep[CORAAL;][]{KendallFarrington2024}.
CORAAL\footnote{CORAAL version 2023.06, last accessed 15 Jan 2026.} is a publicly available collection of sociolinguistic interviews conducted with speakers of AAE.
Concretely, the corpus contains audio recordings and time-aligned transcriptions of over 220 interviews.
The speakers span multiple regions of the United States, such as Detroit (MI), Rochester (NY) and Atlanta (GA), and cover multiple generations, with dates of birth ranging from 1888 to 2005.

Because CORAAL's recordings were made by a range of interviewers, and speakers may shift toward more standardized forms when interviewed by an unfamiliar or non-African American interviewer \citep[cf.\ the observer's paradox;][]{Labov1972}, we restrict our sample to interviews conducted by an African American interviewer with a close relationship to the interviewee, using CORAAL's interviewer metadata.
The 48 interviews meeting both criteria come from four regional subcorpora, viz.\ Washington (DC), Valdosta (GA), Atlanta (GA) and Rochester (NY), with speakers born between 1936 and 1999, and comprise 16,535 conversational turns and 354,412 tokens (20,051 unique).
All subsequent preprocessing and sampling is performed on this subset rather than on CORAAL as a whole.\footnote{All code used for data preprocessing, model prompting and analysis is publicly available at \url{https://github.com/romeodetimmerman/aae-llm-eval}.}

\subsection{Data sampling}
\label{sec:data-sampling}

From the preprocessed interview data (Appendix~\ref{app:preprocessing}), we sample 100 conversational segments consisting of five context turns, where each turn is a single stretch of uninterrupted speech from either the interviewee or the interviewer.
These 500 turns are used as input for both the feature classification task (Section~\ref{sec:task1}) and the next-turn generation task (Section~\ref{sec:task2}).
For the second task, we also pair each segment of five context turns with a sixth subsequent turn, which we refer to as the target turn.
The 100 segments are randomly sampled from the subset discussed above, although a few minimum criteria are considered.
Segments can only be sampled if each turn includes at least five words, to filter out turns which only include backchannel (e.g.\ \lingex{mhm}, \lingex{yeah} or \lingex{ok}).
Additionally, we make sure that no turn appears in more than one segment, so all segments are completely unique. Segments drawn from the same interview nevertheless share speakers and are therefore not fully independent.

\subsection{Model conditions}
\label{sec:model-conditions}

We compare three LLM conditions across both the feature classification and next-turn generation tasks.
The three conditions are Google's Gemini 2.5 Flash \citep{GeminiTeam2025} with reasoning enabled, Gemini 2.5 Flash with reasoning disabled, and OpenAI's GPT-4.1 mini \citep{OpenAI2025GPT41}, which does not support reasoning.
In what follows, we refer to these as Gemini-flash (reasoning), Gemini-flash (no reasoning) and GPT-mini.
While parameter counts are not publicly available for these models, we argue they are comparable in three ways: these models are commercially advertised as general-purpose chatbots, they comprise smaller-tier versions of their respective companies' flagship models, and are released in the same time window, viz.\ April 2025.

These three conditions support two complementary comparisons.
Contrasting the two Gemini-flash conditions against each other isolates the effect of reasoning while holding the rest of the model constant.
Contrasting either Gemini-flash condition against GPT-mini instead varies model family, capturing broader differences in training data, architecture, alignment, etc.\ between model providers.
We use fixed parameters for all three conditions throughout both tasks (i.e.\ temperature = 1.0, top-p = 1.0) to reduce potential confounding when comparing observed differences across the models and prompts under test.
One exception is the reasoning token budget for the Gemini-flash (reasoning) condition, which was converted from dynamic to fixed for the second task because we observed it not spending any tokens on reasoning during testing, as discussed in detail in Section~\ref{sec:t2-generation-setup}.

\section{Task 1: Feature classification}
\label{sec:task1}

The first task focuses on feature classification, and asks to what extent LLMs can correctly detect five AAE morphosyntactic features in conversational excerpts.
We argue that this kind of classification is a precondition for any appropriate downstream handling of AAE.
Building on \citeposs{BlodgettEtAl2020} harms framework introduced earlier, a model that fails to detect AAE features when they occur essentially fails to recognize the existence of the systematic patterns that constitute AAE, which may in turn (re)produce representational or allocational harms.

\subsection{Methods}
\label{sec:t1-methods}

\subsubsection{Features}
\label{sec:features}

We consider five morphosyntactic features of AAE, which we selected based on the sociolinguistic literature on AAE \citep[for an overview, see][]{Green2002}.
Table~\ref{tab:features} describes each feature and gives an example.
Following \citet{MasisEtAl2022}, multiple negation (also called negative concord) can additionally occur in two more specific forms: negative auxiliary inversion, in which a negated auxiliary is fronted to the start of the clause (e.g.\ \lingex{don't nobody know what I had}), and its non-inverted counterpart, in which the subject remains in its usual position (e.g.\ \lingex{nobody don't say nothing}).
Both of these forms, like multiple negation more generally, frequently co-occur with the negative marker \lingex{ain't} (e.g.\ \lingex{I ain't doing nothing wrong}).
For the purposes of this study, we pool these three subtypes of negative concord together.

\begin{table*}[t]
\centering
\small
\begin{tabularx}{\textwidth}{@{}>{\raggedright\arraybackslash}p{0.2\textwidth} >{\raggedright\arraybackslash}X >{\raggedright\arraybackslash}p{0.36\textwidth}@{}}
\toprule
Feature & Description & Example \\
\midrule
Habitual be & An uninflected form of \lingex{to be} marks a habitual or repeated action. & \lingex{he be walking to school} \\
\addlinespace
Copula deletion & A form of \lingex{to be} (i.e.\ the copula) is omitted. & \lingex{she fun} \\
\addlinespace
Third-person singular -s deletion & The -s suffix is absent from a present-tense verb with a third-person singular subject. & \lingex{when he come down here} \\
\addlinespace
Is/was generalization & Singular \lingex{is} or \lingex{was} is used with a plural subject. & \lingex{what they was doing} \\
\addlinespace
Multiple negation & Two or more negative markers within a clause express a single negation. & \lingex{can't nobody tell you it wasn't meant for you} \\
\bottomrule
\end{tabularx}
\caption{The five AAE morphosyntactic features examined in this study, with examples.}
\label{tab:features}
\end{table*}

\subsubsection{Gold annotation}
\label{sec:t1-gold}

The five morphosyntactic features of AAE described above were manually annotated in the 100 conversational segments we sampled from CORAAL (cf.\ Section~\ref{sec:data-sampling}).
The segments were annotated on a turn level, by coding the presence of each of the five features in the 500 context turns, but not in the 100 target turns, since these were specifically selected for Task 2.
A single trained linguist with expertise in the morphosyntax of AAE annotated all turns.

Positive instances are sparse at the turn level.
Across the 500 annotated turns, per-feature counts range from 12 for multiple negation to 32 for copula deletion, with habitual be, third-person -s deletion, and is/was generalization in between at 14, 17 and 31 counts respectively.
Collapsing to the segment level, viz.\ by treating a segment as positive for a feature if any of its five turns is, yields noticeably higher relative frequencies, and modestly lower absolute counts, from 11 to 30 positive segments per feature, since segments rarely contain more than one positive turn for the same feature.
Of the 100 segments, 63 contain at least one of the five features, 22 contain more than one, and 37 contain none.
We consequently treat the segment, rather than the turn, as the unit of analysis for this task.

\subsubsection{Prompting pipeline}
\label{sec:t1-prompting}

For each segment, each model condition (Section~\ref{sec:model-conditions}) predicts whether a given feature occurs anywhere in the five turns.
Each prediction is a JSON object, enforced through structured decoding, consisting of a boolean indicating whether the feature is present and a string quoting the span that triggered a positive judgment, or an empty string when the feature is judged absent.
Models are queried once per feature rather than once for all five features (Appendix~\ref{app:classification-prompts}), and each query is repeated five times, with the five votes aggregated into a single segment-level prediction by majority vote.

We compare two prompting strategies, which we refer to as \textit{baseline} and \textit{self-generated}.
The baseline strategy uses a fixed system prompt per feature, stating the feature's label, definition and an example together with shared annotation guidelines.
These prompts were drafted with a separate LLM outside the conditions under study (i.e.\ Claude Sonnet 5) and finalized by hand (Appendix~\ref{app:baseline-prompts}).
This reflects a human-in-the-loop, cross-model process typical of hand-engineered prompting.

The self-generated strategy instead has each model condition write its own system prompt per feature, without human editing.
We use a single meta prompt, drafted by Claude Sonnet 5 and edited by hand (Appendix~\ref{app:selfgen-prompts}), to generate these prompts once per model and feature, which are then reused for all predictions.
This strategy may reveal the potential upside of letting a model partly decide how it should be instructed.
Since the baseline strategy produces high false positive rates across conditions (Section~\ref{sec:t1-overall}), the meta prompt explicitly tells each model that its baseline judgments produce too many false positives and asks it to prioritize precision.

\begin{table*}[t]
\centering
\small
\begin{tabular}{l c c c}
\toprule
Condition & Precision & Recall & F1 \\
\midrule
Gemini-flash (reasoning), baseline & 0.74 [0.66, 0.81] & \textbf{0.99} [0.96, 1.00] & 0.84 [0.78, 0.89] \\
Gemini-flash (reasoning), self-gen. & \textbf{0.85} [0.77, 0.90] & 0.98 [0.90, 1.00] & \textbf{0.91} [0.86, 0.94] \\
Gemini-flash (no reasoning), baseline & 0.49 [0.42, 0.56] & 0.95 [0.90, 0.98] & 0.62 [0.56, 0.69] \\
Gemini-flash (no reasoning), self-gen. & 0.57 [0.49, 0.65] & 0.91 [0.84, 0.95] & 0.69 [0.62, 0.76] \\
GPT-mini, baseline & 0.51 [0.43, 0.59] & 0.92 [0.86, 0.96] & 0.62 [0.56, 0.70] \\
GPT-mini, self-gen. & 0.61 [0.50, 0.71] & 0.75 [0.66, 0.81] & 0.65 [0.57, 0.72] \\
\citet{MasisEtAl2022} & 0.78 [0.67, 0.86] & 0.70 [0.61, 0.78] & 0.73 [0.65, 0.80] \\
\bottomrule
\end{tabular}
\caption{Task 1 (feature classification): macro-averaged precision, recall and F1 across the five AAE features, for each LLM condition under both prompting strategies and for the supervised \citet{MasisEtAl2022} classifier, computed over the 100 segments with 95\% bootstrap confidence intervals. The highest value per metric is shown in bold.}
\label{tab:t1-macro-performance}
\end{table*}

\begin{figure*}[t]
\centering
\includegraphics[width=0.8\textwidth]{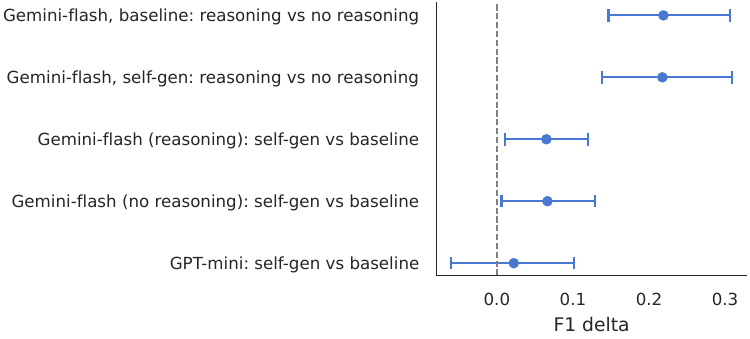}
\caption{Point estimates and 99\% family-wise confidence intervals for the five pre-specified macro F1 contrasts. An interval crossing zero (dashed line) marks a difference that is not significant after Bonferroni correction ($m = 5$).}
\label{fig:t1-f1-contrasts}
\end{figure*}

\subsubsection{Evaluation}
\label{sec:t1-evaluation}

\paragraph{Performance metrics.}
To measure overall performance, we compare, for each feature and model condition, a segment's predicted label against its gold label to obtain segment-level true positives (TP), false positives (FP), true negatives (TN) and false negatives (FN).
From these counts, we compute precision, recall and F1 separately for each of the five features, and then average these five per-feature scores into single macro precision, recall and F1 metrics per condition.

\paragraph{Bootstrap procedure.}
To quantify uncertainty in the per-condition performance metrics defined above, we use a nonparametric bootstrap method, following \citeposs{Miller2024} case for rigorous statistical treatment of LLM evaluation results.
Rather than resampling feature-level decisions, we instead resample at the level of the segment (cf.\ supra).
We consequently draw bootstrap samples of 100 segments with replacement over $B = 10{,}000$ iterations, applying the same resampled set of segments to every condition on a given iteration so that comparisons across conditions remain paired.
Macro precision, recall and F1 are then recomputed on these resampled data.

We compute two kinds of confidence interval on the basis of this bootstrap procedure.
First, we report marginal 95\% confidence intervals for each condition's macro and per-feature precision, recall and F1.
We use a bias-corrected and accelerated method \citep[i.e.\ BCa method;][]{EfronTibshirani1998,DiCiccioEfron}, which accounts for skewness as these metrics approach their ceiling value of 1.
These marginal intervals are mostly descriptive, and we run no significance tests at the per-feature level.
Second, we test a pre-specified set of paired contrasts on F1, which we consider to be the primary metric for this task.
These contrasts are Gemini with versus without reasoning under each of the two prompting pipelines (two contrasts), and the self-generated versus baseline prompt for each of the three model conditions (three contrasts).
We treat these contrasts as one Bonferroni family ($m = 5$) and control the family-wise error rate at $\alpha = 0.05$.\footnote{We use Bonferroni correction here for its straightforward control of the family-wise error rate, though we note its conservative nature may be a limitation given our modest sample sizes.}

\begin{figure*}[t]
\centering
\includegraphics[width=0.8\textwidth]{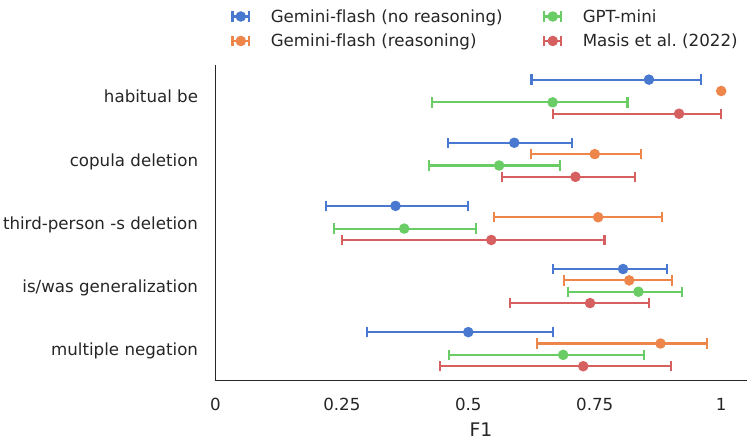}
\caption{Per-feature F1 with 95\% confidence intervals for the three LLM conditions under the baseline prompt and the \citet{MasisEtAl2022} classifier.}
\label{fig:t1-per-feature-f1}
\end{figure*}

\subsubsection{Supervised baseline}
\label{sec:t1-supervised-baseline}

We additionally compare the three model conditions described in Section~\ref{sec:data-and-models} against a supervised baseline, viz.\ a task-specific classifier developed by \citet{MasisEtAl2022}.
Their classifier comprises a shared BERT encoder, finetuned on CORAAL data, with independent binary classification heads for 17 morphosyntactic features of AAE.
We reuse Masis et al.'s publicly released trained weights, and repurpose seven of their 17 heads to match the features examined in this study (Appendix~\ref{app:supervised-baseline}).
Since this classifier was finetuned on CORAAL, we consider its scores to be in-domain reference results to compare against the zero-shot LLM conditions.

\subsection{Results}
\label{sec:t1-results}

\subsubsection{Overall performance}
\label{sec:t1-overall}

Table~\ref{tab:t1-macro-performance} reports macro-averaged precision, recall and F1 for the three LLM conditions under both prompting strategies, alongside the \citet{MasisEtAl2022} classifier. 
Gemini-flash (reasoning) is the clear top performer under either prompting strategy, with a macro F1 of 0.91 under the self-generated prompt and 0.84 under the baseline prompt.
The \citet{MasisEtAl2022} classifier (0.73) falls below Gemini-flash (reasoning) but outperforms the other two LLM conditions, Gemini-flash (no reasoning) and GPT-mini, which perform noticeably worse (0.62--0.69).

The table also reveals a distinct precision/recall asymmetry. 
Under the baseline prompt, all three LLM conditions achieve very high recall (0.92--0.99) but considerably lower precision (0.49--0.74). 
The BERT-style classifier, on the other hand, shows the opposite pattern, with the lowest recall of any condition (0.70) but the second-highest precision (0.78).
The self-generated prompt, designed to address this imbalance, raises precision for all three LLM conditions (+0.08 to +0.11), but its recall cost differs by model.
For both Gemini-flash conditions, the recall drop is small (-0.01 and -0.04), so precision gains translate into higher F1. For GPT-mini, however, the drop is large (-0.17), leaving its F1 roughly unchanged (+0.02).

Testing these differences through five pre-specified contrasts (Figure~\ref{fig:t1-f1-contrasts}) shows that the self-generated prompt yields a small but significant F1 gain for Gemini-flash (reasoning) (+0.065 [0.011, 0.120]) and Gemini-flash (no reasoning) (+0.066 [0.006, 0.129]), but not for GPT-mini (+0.022 [-0.061, 0.101]).
Reasoning, by contrast, has a large and significant effect, raising Gemini-flash's macro F1 by +0.219 [0.147, 0.307] under the baseline prompt and a quasi-identical +0.218 [0.138, 0.310] under the self-generated prompt.
Appendix~\ref{app:pairwise-tests} gives the full test results, including Bonferroni-adjusted p-values.

\subsubsection{Per-feature performance}
\label{sec:t1-per-feature}

Figure~\ref{fig:t1-per-feature-f1} reports per-feature F1 with 95\% confidence intervals under the baseline prompt, showing that no single feature is uniformly the hardest to classify.
Instead, performance depends on the interaction between feature and condition.
Similarly, the reasoning advantage described above is not uniform across all features.
Gemini-flash (reasoning) improves over its non-reasoning counterpart most on multiple negation (F1 0.50 versus 0.88) and third-person -s deletion (0.36 versus 0.76), while the two are essentially tied on is/was generalization (0.81 versus 0.82).
Is/was generalization is the one feature every condition handles comparably well (0.74--0.84), and the only one on which GPT-mini matches Gemini-flash (reasoning).

Since F1 masks the precision/recall asymmetry noted above, Figure~\ref{fig:t1-per-feature-precision-recall} plots per-feature precision against recall (full breakdown in Appendix~\ref{app:per-feature-table}).
The LLM conditions detect almost every gold-positive segment (recall is at its ceiling of 1.0 for habitual be and multiple negation under all three conditions, and high elsewhere), so their per-feature F1 is driven almost entirely by precision.
The low F1s for third-person -s deletion and multiple negation are hence precision shortcomings: both non-reasoning LLMs heavily overpredict third-person -s deletion (precision 0.22 and 0.23), as does Gemini-flash (no reasoning) for multiple negation (0.33).
The \citet{MasisEtAl2022} classifier, conversely, struggles most to recover third-person -s deletion (recall 0.43).
An analysis of the spans quoted in positive predictions (Appendix~\ref{sec:t1-span-validation}) shows that these false positives are almost entirely misjudgments of genuine speech rather than hallucinated evidence. 
Only about 4\% of false-positive spans did not match the transcript, and these were mostly non-verbatim renderings of real speech.
For true positives, models localize features carrying a lexical element (e.g.\ multiple negation) almost perfectly, but are less accurate for absence-based features such as copula deletion.

\begin{figure}[t]
\centering
\includegraphics[width=\columnwidth]{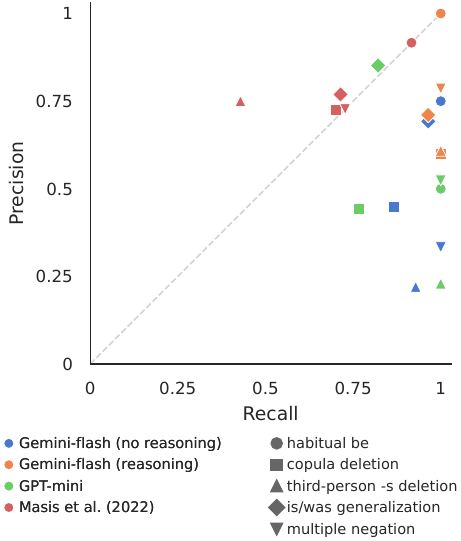}
\caption{Per-feature precision against recall for the three LLM conditions under the baseline prompt and the \citet{MasisEtAl2022} classifier. Color shows model condition while shape shows AAE feature.}
\label{fig:t1-per-feature-precision-recall}
\end{figure}

\section{Task 2: Next-turn generation}
\label{sec:task2}

While the first task gauges whether models can recognize AAE, we argue it is equally important to examine how models respond to AAE input, since a model could recognize AAE features without producing them, or vice versa.
The second task consequently focuses on next-turn generation, and asks to what extent LLMs generate appropriate and diverse dialogue turns when given conversational AAE input as context.
Even if models do use features of AAE during generation, this may not necessarily result in desirable output \citep[cf.][]{BasoahEtAl2025,FleisigEtAl2024}, so we additionally examine the pragmatic quality of the generated turns, i.e.\ whether a turn is a natural, coherent and contextually appropriate contribution to the ongoing conversation.
We also explore how diverse generations are across repeated attempts at the same continuation, verifying if model-generated turns tend to collapse into similar, templated or even stereotyped responses.

\subsection{Methods}
\label{sec:t2-methods}

\subsubsection{Conversational context}
\label{sec:t2-context}

We build on the same data used for the first task, viz.\ the 100 conversational segments of five speaker turns described in Section~\ref{sec:data-sampling}.
For this task, those five turns serve as the conversational context a model is asked to continue.
Each segment is presented to the models as the five context turns in their original order, one turn per line, with each line consisting of the anonymized speaker label and that speaker's utterance.
The model is additionally told explicitly which of the two speakers utters the next turn, since the interviewer and interviewee take up different roles in the interaction.

Alongside the context and the next speaker, the model receives a one-line description of what the next turn should convey.
For this, we use a sixth turn for each of the 100 segments, i.e.\ the target turn (cf.\ Section~\ref{sec:data-sampling}).
This target turn is never shown to the models.
Instead, we show summaries following a template of \lingex{in this turn, the speaker\ldots} (e.g.\ \lingex{in this turn, the speaker mentions they don't know how their family got their first home but recalls living in Bladensburg}).
Without such a description, the task would be open-ended in a way that makes the generations rather uninterpretable for our purposes, since we are mostly interested in differences in the linguistic form, rather than the content.
The summaries were drafted automatically from the withheld target turns by a separate LLM outside the three conditions under study (i.e.\ Claude Sonnet 5, as with the classification prompts in Section~\ref{sec:t1-prompting}), and were subsequently reviewed and corrected by hand.

\begin{figure*}[t]
\centering
\includegraphics[width=0.8\textwidth]{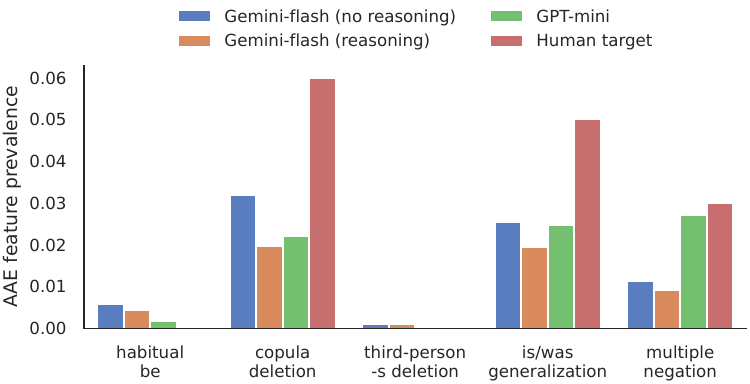}
\caption{Proportion of generations in which the \citet{MasisEtAl2022} classifier detects each feature, computed over the 5,000 generations per condition, with the rate in the 100 human target turns shown as a reference.}
\label{fig:t2-aae-prevalence}
\end{figure*}

\subsubsection{Generation setup}
\label{sec:t2-generation-setup}

We use the same system prompt for every segment and every model condition, instructing the model to continue a transcript of a naturally occurring conversational dialogue from a recorded interview, in which two speakers take turns, and asking it to produce the next turn that the indicated speaker would most naturally say at this point (cf.\ Appendix~\ref{app:t2-prompt}).
Two properties of this system prompt are noteworthy.
First, it frames the task as the continuation of a spoken transcript rather than as a chat exchange, since a chatbot framing may invite the register these models are aligned to produce by their providers.
Second, the prompt is entirely dialect-neutral, since it makes no mention of AAE or MAE, nor does it use the terms dialect, variety or register.
We similarly give no example continuation turns.
In other words, any AAE feature appearing in a generation is a spontaneous response to the conversational context, emulating a speaker's accommodation to their interlocutor.

We generate 50 turns per segment for each of the three model conditions described in Section~\ref{sec:model-conditions}, yielding 5,000 generations per condition and 15,000 in total, allowing us to analyze the diversity in a model's generated next turns for the same conversational segment.
All conditions use the same parameters as in the first task, viz.\ a temperature of 1.0 and a top-p of 1.0, except for one, viz.\ the reasoning budget.
For the first task, we gave the Gemini-flash (reasoning) condition a dynamic reasoning budget, but when pilot testing the present task, a dynamic setting resulted in no reasoning tokens being spent at all on producing the next turns.
Since this would collapse the Gemini-flash (reasoning) and Gemini-flash (no reasoning) conditions, we fixed the reasoning condition's budget at 1,024 tokens, which was selected empirically after comparing budgets of 512 and 1,024 tokens.

\subsubsection{Evaluation}
\label{sec:t2-evaluation}

\paragraph{AAE feature prevalence.}
To measure to what extent the models produce AAE, we classify every generation for the potential presence of the same five morphosyntactic features examined in the first task.
We use the supervised classifier of \citet{MasisEtAl2022} introduced in Section~\ref{sec:t1-supervised-baseline} for this part of the evaluation.
While the Gemini-flash (reasoning) condition clearly outperformed the Masis et al.\ classifier on the first task, we nevertheless prefer the BERT-style model here, since it is independent of all three generators and shares no architecture or provider with them.
In other words, it should not systematically favor the outputs of one condition over another.
Prevalence is reported as the proportion of generations in which a feature is detected, computed per condition and per feature, alongside the corresponding rate in the human target turns.
Note that this comparison rests on only 100 turns (one per segment), so generation versus target-turn differences should be read with some caution.

\paragraph{Diversity.}
To approximate how wide a model's distribution of plausible continuations is, we compute Self-BLEU \citep{ZhuEtAl2018} over the 50 generations produced for each segment by each model condition.
Self-BLEU adapts BLEU, a metric originally developed for machine translation that scores a text by the proportion of its n-grams that also occur in one or more reference texts, by using a model's own generations as references for one another.
In casu, each generation is scored against the other 49 as references, and the resulting scores are averaged within a segment and then across segments, resulting in a single metric per condition.
A higher Self-BLEU means the generations resemble one another more closely, and therefore that the condition is less diverse.

\paragraph{Pragmatic quality.}
To assess pragmatic quality, we conduct a blinded pairwise preference study over 200 trials, all judged by a single rater, viz.\ the same trained linguist who produced the gold annotations for the first task (Section~\ref{sec:t1-gold}).
Each trial pairs one of the 100 context segments with one randomly selected generation from each of two randomly drawn model conditions, shown in random left/right order alongside the five context turns and the target-turn summary, but not the target turn itself.
The rater prefers one generation or declares a tie, based on the turn's overall pragmatic quality and, in particular, its plausibility as a next turn in the conversation.

\subsection{Results}
\label{sec:t2-results}

\subsubsection{AAE feature prevalence}
\label{sec:t2-prevalence}

Figure~\ref{fig:t2-aae-prevalence} shows the proportion of generations in which the \citet{MasisEtAl2022} classifier detects each of the five features, computed over the 5,000 generations per condition, alongside the corresponding rate in the 100 human target turns as a reference.

As Figure~\ref{fig:t2-aae-prevalence} shows, AAE production is low across the board.
At least one of the five features is detected in 7.1\% of the generations for Gemini-flash (no reasoning) and 7.5\% for GPT-mini, against 5.0\% for Gemini-flash (reasoning), while 14\% of the human target turns carry at least one feature.
On a per-feature level, every condition falls below the human reference on copula deletion (0.020--0.032 versus 0.060), is/was generalization (0.019--0.025 versus 0.050) and multiple negation (0.009--0.027 versus 0.030).
Habitual be and third-person -s deletion are virtually absent from the generations (rates of at most 0.006 and 0.001 respectively), but they are equally absent from the target turn.

The ordering across conditions is fairly consistent.
Gemini-flash (reasoning) produces the least AAE overall, and has the lowest rate of the three on copula deletion, is/was generalization and multiple negation.
In other words, reasoning, which clearly improved feature classification in the first task, may push generation toward a more MAE-aligned next turn.

We also observe that the detected presence of features is distributed unevenly across the segments.
Only 56 of the 100 segments contain any detected AAE at all for Gemini-flash (no reasoning), against 39 for GPT-mini and 37 for Gemini-flash (reasoning).
Within the segments that do contain some AAE, the median number of generations carrying a feature is 4 out of 50 for Gemini-flash (no reasoning) and GPT-mini, and 2 out of 50 for Gemini-flash (reasoning), while a handful of segments reach 46 to 49.
AAE consequently surfaces in concentrated bursts, viz.\ in a small set of contexts where a condition produces it repeatedly, rather than thinly across the sample as a whole.

\subsubsection{Generation diversity}
\label{sec:t2-diversity}

\begin{figure}[t]
\centering
\includegraphics[width=\columnwidth]{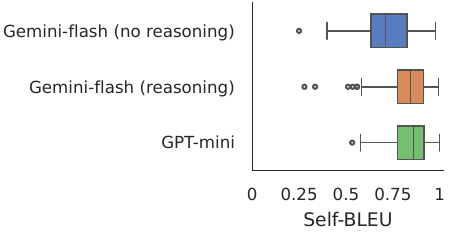}
\caption{Distribution of segment-level Self-BLEU across the 100 segments, per condition.}
\label{fig:t2-self-bleu}
\end{figure}

As Figure~\ref{fig:t2-self-bleu} shows, segment-level Self-BLEU is high for all three conditions.
It is the highest for GPT-mini (mean 0.84, SD 0.10), closely followed by Gemini-flash (reasoning) (mean 0.82, SD 0.13), while Gemini-flash (no reasoning) falls somewhat lower (mean 0.71, SD 0.15).
We hence conclude that, across the model conditions, the fifty attempts at the same continuation share the bulk of their n-grams.
Of the three conditions, Gemini-flash (no reasoning) is the most varied, which suggests that enabling reasoning may decrease the output diversity.
This is consistent with the pattern described above, where enabling reasoning appears to correlate with a decrease in AAE feature production.

It is worth acknowledging that part of the n-gram overlap revealed here occurs by construction, since every generation for a segment is conditioned on the same one-line summary of the target turn (cf.\ Section~\ref{sec:t2-context}).
The comparison between conditions nevertheless remains informative, however, as all three operate under this same constraint.
Note also that GPT-mini's generated turns are noticeably longer on average (mean length of 24 words, versus 16 and 18 for the two Gemini-flash conditions), which we will return to in the next section.

\subsubsection{Pragmatic quality}
\label{sec:t2-quality}

Table~\ref{tab:t2-win-rate} shows the raw winrate of each condition against each other condition, with ties counted as half of a win to either side.
The rater declares a tie in 62 of the 200 trials (31\%), leaving 138 decisive judgments.
As the table shows, GPT-mini is preferred to both Gemini-flash conditions by a wide margin, winning 76\% of its trials against Gemini-flash (no reasoning) and 72\% against Gemini-flash (reasoning).
The two Gemini-flash conditions, on the other hand, are closer together, with Gemini-flash (reasoning) winning 61\% of their shared trials.

\begin{table}[t]
\centering
\resizebox{\columnwidth}{!}{%
\begin{tabular}{l c c c}
\toprule
& \makecell{Gemini-flash\\(no reasoning)} & \makecell{Gemini-flash\\(reasoning)} & GPT-mini \\
\midrule
Gemini-flash (no reasoning) & -- & 0.39 & 0.24 \\
Gemini-flash (reasoning) & 0.61 & -- & 0.28 \\
GPT-mini & 0.76 & 0.72 & -- \\
\bottomrule
\end{tabular}}
\caption{Pairwise winrate matrix.}
\label{tab:t2-win-rate}
\end{table}

Finally, we formally test the pairwise differences between the model conditions through an exact binomial sign test over each pair's non-tied trials, treating the three tests as a single Bonferroni family (Appendix~\ref{app:pragmatic-quality}).
Both contrasts involving GPT-mini are robust, since it wins 38 of the 46 decisive trials against Gemini-flash (no reasoning) and 42 of the 53 against Gemini-flash (reasoning) (Bonferroni-adjusted $p < .001$ in both cases).
The contrast between the two Gemini-flash conditions, on the other hand, does not reach the family-wise threshold (Bonferroni-adjusted $p = .071$).
It is worth noting that 31 of 70 trials (44\%) for this pair end in a tie, against 21\% and 26\% for the two pairs involving GPT-mini.
We hence tentatively conclude that reasoning improves the pragmatic quality of a next turn only modestly, considerably less than it improved feature classification.

An additional descriptive analysis shows that turn length may be a confounding variable here.
Across the 138 decisive trials, the preferred turn averages 24.1 words against 19.2 for the rejected one, and the longer of the two wins 72\% of the time.
Since GPT-mini's generations are the longest of the three conditions on average (cf.\ supra), part of its advantage may reflect a preference for more elaborated turns, which the rater may have read as more conversational.

\section{Discussion and conclusion}
\label{sec:discussion}
\label{sec:conclusion}

When bringing the results of both tasks together, the most striking pattern is the differing performance of the Gemini-flash (reasoning) condition.
It is the strongest detector by a clear margin, outperforming all other conditions considerably.
On the generation task, however, it produces AAE features the least often of the three conditions, both overall and on a feature-level, and it is also markedly less diverse than its non-reasoning counterpart, generating the least varied and least unique set of continuations of any condition.
Its performance on the human preference task shows that it is marginally preferred over its non-reasoning counterpart, but decisively outperformed by GPT-mini.
From this pattern, we conclude that a model's capability to recognize and classify AAE features does not straightforwardly translate into its spontaneous accommodation to AAE speech as conversational input.

We connect these findings to the existing literature on linguistic bias in NLP introduced in Section~\ref{sec:background}, focusing particularly on representational harms, which \citet{BlodgettEtAl2020} describe as arising when a system represents a social group less favorably than others, demeans it, or fails to recognize its existence altogether.
Our results are most relevant to this last element, viz.\ the recognition of the linguistic variation associated with a particular social group.
On the one hand, the Task 1 findings suggest that the model conditions examined here, and Gemini-flash (reasoning) in particular, are able to recognize linguistic patterns associated with AAE with some reliability.
On the other hand, the Task 2 findings point in a different direction, since models overwhelmingly respond in a register aligned with MAE.
This asymmetry suggests that a model's capacity to recognize AAE does not, on its own, translate into equitable treatment of AAE-speaking users.

It also raises a question about the potential role of prompting.
Task 1 explicitly prompts each model to detect a given feature.
Moreover, our results show models to be sensitive to how that prompt is engineered, since performance predominantly improved when models were given a self-generated prompt with more specific precision-oriented instructions.
Task 2, by contrast, includes no mention of dialect, variety or register at all.
Instead, models are simply asked to produce a plausible conversational continuation, so that any accommodation to AAE, or lack thereof, reflects a spontaneous response to context.
One might therefore conclude that a potential solution is to prompt models more explicitly to reply in the same variety as their interlocutor.
However, as \citet{FleisigEtAl2024} document, when LLMs are explicitly asked to imitate the writing style of a particular linguistic variety, including AAE, their responses show reduced comprehension of the input and are especially prone to stereotyping.

More conceptually, even if explicit dialect-prompting could be engineered to avoid the pitfalls \citet{FleisigEtAl2024} document, we would not consider it a desirable solution.
What matters, in our view, is a model's capacity to attend and respond to context spontaneously, without being specifically instructed to do so, since requiring speakers of a particular linguistic variety to explicitly request accommodation risks placing the burden of securing equitable treatment on the very social groups already disadvantaged by these systems.
We consequently approach the performance trends observed in this study less as a matter of prompt design, and more as pointing to the data and alignment processes further upstream, such as AAE's marginal and often unrepresentative presence in pretraining corpora \citep{DeasEtAl2025}, and evidence that the reward models used to align these systems specifically disprefer AAE-aligned responses and steer generation toward MAE, even when prompted in AAE \citep{MireEtAl2025}.
Additionally, our results for Gemini-flash (reasoning) suggest that reasoning may also push generation away from AAE, possibly because reasoning-oriented training data, which typically centers on mathematical, logical or coding problems, may rely on standardized registers, a hypothesis consistent with this condition's lower AAE feature prevalence and lower output diversity relative to its non-reasoning counterpart.
We believe this calls for further research into both (i) how these processes reproduce dialectal bias, and (ii) how future pre- and posttraining pipelines might instead be designed to represent and accommodate dialectal variation more equitably, potentially leveraging authentic, naturalistic language resources such as CORAAL \citep{KendallFarrington2024}.

\section*{Limitations}

We acknowledge several limitations to this study.
A first limitation is that both the gold morphosyntactic feature labels used to evaluate the classification task (cf.\ Section~\ref{sec:t1-gold}) and the 200 pairwise preference judgments used to assess generation quality (cf.\ Section~\ref{sec:t2-evaluation}) were produced by a single trained rater, with no second annotator or inter-rater reliability check.
For the former, we attempted to mitigate this concern by specifically selecting five features of AAE which are well-defined morphosyntactic categories with established annotation criteria.
For the latter, however, the limitation may be more substantial, since these pairwise preference decisions are, by design, built on a subjective notion of pragmatic quality.
Moreover, no AAE-speaking annotators were available for this study, so neither the gold labels nor the preference judgments reflect native-speaker intuitions, which may differ, particularly for pragmatic quality.
Future work could therefore replicate parts of this study with multiple annotators who are speakers of AAE.

A second limitation concerns the generalizability of the findings presented here.
As discussed in Section~\ref{sec:t1-gold}, the sample of 100 segments yields sparse positive counts for some features at the turn level (e.g.\ 12 turns for multiple negation), so per-feature estimates carry wide uncertainty.
Moreover, the study only investigated five morphosyntactic features of AAE.
Future work could examine more features, and, more importantly, also other dimensions of linguistic variation, such as lexical, pragmatic or even phonological features, since many phonological patterns may be expressed orthographically \citep[cf.][]{ZiemsEtAl2022,DeasEtAl2024}.

A third limitation involves the model conditions examined in this study.
As highlighted in Section~\ref{sec:model-conditions}, Gemini 2.5 Flash and GPT-4.1 mini are both smaller-tier alternatives to their providers' flagship models.
Moreover, they are closed-source models, so their parameter counts, training data and architecture are not publicly documented.
Future work could extend the model conditions examined here to flagship-tier and/or open-weight models.

A final, more conceptual limitation concerns the study's framing of detection and generation as separate tasks.
In actual model inference, interpreting conversational context and generating a response are not cleanly separable, since an LLM attends to context and autoregressively produces output within a single forward pass.
Although we consider this more so an open question for the behavioral evaluation of LLMs more generally rather than a limitation specific to this study, it is worth pointing out the difficulty in disentangling LLM comprehension and production, especially when operating within the bounds of closed-source, API-based methodologies.

\bibliography{references}

\begin{thebibliography}{36}
\providecommand{\natexlab}[1]{#1}

\bibitem[{Basoah et~al.(2025)Basoah, Cunningham, Adams, Bose, Jain, Yadav,
  Yang, Reinecke, and Rosner}]{BasoahEtAl2025}
Jeffrey Basoah, Jay~L. Cunningham, Erica Adams, Alisha Bose, Aditi Jain,
  Kaustubh Yadav, Zhengyang Yang, Katharina Reinecke, and Daniela Rosner. 2025.
\newblock \href {https://doi.org/10.1145/3757423} {Should {{AI Mimic People}}?
  {{Understanding AI-Supported Writing Technology Among Black Users}}}.
\newblock \emph{Preprint}, arXiv:2505.00821.

\bibitem[{Blodgett et~al.(2020)Blodgett, Barocas, Daum{\'e}~Iii, and
  Wallach}]{BlodgettEtAl2020}
Su~Lin Blodgett, Solon Barocas, Hal Daum{\'e}~Iii, and Hanna Wallach. 2020.
\newblock \href {https://doi.org/10.18653/v1/2020.acl-main.485} {Language
  ({{Technology}}) is {{Power}}: {{A Critical Survey}} of ``{{Bias}}'' in
  {{NLP}}}.
\newblock In \emph{Proceedings of the 58th {{Annual Meeting}} of the
  {{Association}} for {{Computational Linguistics}}}, pages 5454--5476, Online.
  Association for Computational Linguistics.

\bibitem[{Blodgett and O'Connor(2017)}]{BlodgettOConnor2017}
Su~Lin Blodgett and Brendan O'Connor. 2017.
\newblock \href {https://doi.org/10.48550/arXiv.1707.00061} {Racial
  {{Disparity}} in {{Natural Language Processing}}: {{A Case Study}} of
  {{Social Media African-American English}}}.
\newblock \emph{Preprint}, arXiv:1707.00061.

\bibitem[{Blodgett et~al.(2018)Blodgett, Wei, and O'Connor}]{BlodgettEtAl2018}
Su~Lin Blodgett, Johnny Wei, and Brendan O'Connor. 2018.
\newblock \href {https://doi.org/10.18653/v1/P18-1131} {Twitter {{Universal
  Dependency Parsing}} for {{African-American}} and {{Mainstream American
  English}}}.
\newblock In \emph{Proceedings of the 56th {{Annual Meeting}} of the
  {{Association}} for {{Computational Linguistics}} ({{Volume}} 1: {{Long
  Papers}})}, pages 1415--1425, Melbourne, Australia. Association for
  Computational Linguistics.

\bibitem[{Bradley and Terry(1952)}]{BradleyTerry1952}
Ralph~Allan Bradley and Milton~E. Terry. 1952.
\newblock \href {https://doi.org/10.2307/2334029} {Rank {{Analysis}} of
  {{Incomplete Block Designs}}: {{I}}. {{The Method}} of {{Paired
  Comparisons}}}.
\newblock \emph{Biometrika}, 39(3/4):324--345.

\bibitem[{Charity et~al.(2004)Charity, Scarborough, and
  Griffin}]{CharityEtAl2004}
Anne~H. Charity, Hollis~S. Scarborough, and Darion~M. Griffin. 2004.
\newblock \href {https://doi.org/10.1111/j.1467-8624.2004.00744.x} {Familiarity
  with {{School English}} in {{African American}} children and its relation to
  early reading achievement}.
\newblock \emph{Child Development}, 75(5):1340--1356.

\bibitem[{Dacon(2022)}]{Dacon2022}
Jamell Dacon. 2022.
\newblock \href {https://doi.org/10.48550/ARXIV.2206.08978} {Towards a {{Deep
  Multi}}-layered {{Dialectal Language Analysis}}: {{A Case Study}} of
  {{African-American English}}}.
\newblock \emph{arXiv preprint}.

\bibitem[{Deas et~al.(2024)Deas, Grieser, Hou, Kleiner, Martin, Nandanampati,
  Patton, and McKeown}]{DeasEtAl2024}
Nicholas Deas, Jessi Grieser, Xinmeng Hou, Shana Kleiner, Tajh Martin, Sreya
  Nandanampati, Desmond Patton, and Kathleen McKeown. 2024.
\newblock {{PhonATe}}: {{Impact}} of {{Type-Written Phonological Features}} of
  {{African American Language}} on {{Generative Language Modeling Tasks}}.
\newblock In \emph{First {{Conference}} on {{Language Modeling}} ({{COLM}}
  2024)}.

\bibitem[{Deas et~al.(2023)Deas, Grieser, Kleiner, Patton, Turcan, and
  McKeown}]{DeasEtAl2023}
Nicholas Deas, Jessi Grieser, Shana Kleiner, Desmond Patton, Elsbeth Turcan,
  and Kathleen McKeown. 2023.
\newblock \href {https://doi.org/10.48550/arXiv.2305.14291} {Evaluation of
  {{African American Language Bias}} in {{Natural Language Generation}}}.
\newblock \emph{Preprint}, arXiv:2305.14291.

\bibitem[{Deas et~al.(2025)Deas, Vente, Ananthram, Grieser, Patton, Kleiner,
  Shepard, and McKeown}]{DeasEtAl2025}
Nicholas Deas, Blake Vente, Amith Ananthram, Jessica~A. Grieser, Desmond
  Patton, Shana Kleiner, James Shepard, and Kathleen McKeown. 2025.
\newblock \href {https://doi.org/10.48550/arXiv.2503.10789} {Data
  {{Caricatures}}: {{On}} the {{Representation}} of {{African American
  Language}} in {{Pretraining Corpora}}}.
\newblock \emph{Preprint}, arXiv:2503.10789.

\bibitem[{DiCiccio and Efron(1996)}]{DiCiccioEfron}
Thomas~J DiCiccio and Bradley Efron. 1996.
\newblock Bootstrap {{Confidence Intervals}}.
\newblock \emph{Statistical Science}, 11(3).

\bibitem[{Efron and Tibshirani(1998)}]{EfronTibshirani1998}
Bradley Efron and Robert Tibshirani. 1998.
\newblock \emph{An Introduction to the Bootstrap}, nachdr. edition.
\newblock Number~57 in Monographs on Statistics and Applied Probability.
  Chapman \& Hall, Boca Raton, Fla.

\bibitem[{Finch et~al.(2025)Finch, Paek, Choi, and Choi}]{FinchEtAl2025}
Sarah~E. Finch, Ellie~S. Paek, Ikseon Choi, and Jinho~D. Choi. 2025.
\newblock \href {https://doi.org/10.48550/arXiv.2501.03441} {Finding {{A
  Voice}}: {{Exploring}} the {{Potential}} of {{African American Dialect}} and
  {{Voice Generation}} for {{Chatbots}}}.
\newblock \emph{Preprint}, arXiv:2501.03441.

\bibitem[{Fleisig et~al.(2024)Fleisig, Smith, Bossi, Rustagi, Yin, and
  Klein}]{FleisigEtAl2024}
Eve Fleisig, Genevieve Smith, Madeline Bossi, Ishita Rustagi, Xavier Yin, and
  Dan Klein. 2024.
\newblock \href {https://doi.org/10.48550/ARXIV.2406.08818} {Linguistic
  {{Bias}} in {{ChatGPT}}: {{Language Models Reinforce Dialect
  Discrimination}}}.
\newblock \emph{arXiv preprint}.

\bibitem[{{Gemini Team}(2025)}]{GeminiTeam2025}
{Gemini Team}. 2025.
\newblock \href {https://arxiv.org/abs/2507.06261} {Gemini 2.5: {{Pushing}} the
  frontier with advanced reasoning, multimodality, long context, and next
  generation agentic capabilities}.
\newblock \emph{Preprint}, arXiv:2507.06261.

\bibitem[{Graves et~al.(2024)Graves, Aswar, Desai, Nampelli, Chakraborty, and
  Hall}]{GravesEtAl2024}
Eric Graves, Shreyas Aswar, Rujuta Desai, Srilekha Nampelli, Sunandan
  Chakraborty, and Ted Hall. 2024.
\newblock \href {https://doi.org/10.1145/3674829.3675060} {{{AAVE Corpus
  Generation}} and {{Low-Resource Dialect Machine Translation}}}.
\newblock In \emph{Proceedings of the 7th {{ACM SIGCAS}}/{{SIGCHI Conference}}
  on {{Computing}} and {{Sustainable Societies}}}, pages 50--59, New Delhi
  India. ACM.

\bibitem[{Green(2002)}]{Green2002}
Lisa~J. Green. 2002.
\newblock \href {https://doi.org/10.1017/CBO9780511800306} {\emph{African
  {{American English}}: {{A Linguistic Introduction}}}}, 1 edition.
\newblock Cambridge University Press.

\bibitem[{Groenwold et~al.(2020)Groenwold, Ou, Parekh, Honnavalli, Levy, Mirza,
  and Wang}]{GroenwoldEtAl2020}
Sophie Groenwold, Lily Ou, Aesha Parekh, Samhita Honnavalli, Sharon Levy, Diba
  Mirza, and William~Yang Wang. 2020.
\newblock \href {https://doi.org/10.48550/arXiv.2010.02510} {Investigating
  {{African-American Vernacular English}} in {{Transformer-Based Text
  Generation}}}.
\newblock \emph{Preprint}, arXiv:2010.02510.

\bibitem[{Harris et~al.(2022)Harris, Halevy, Howard, Bruckman, and
  Yang}]{HarrisEtAl2022}
Camille Harris, Matan Halevy, Ayanna Howard, Amy Bruckman, and Diyi Yang. 2022.
\newblock \href {https://doi.org/10.1145/3531146.3533144} {Exploring the
  {{Role}} of {{Grammar}} and {{Word Choice}} in {{Bias Toward African American
  English}} ({{AAE}}) in {{Hate Speech Classification}}}.
\newblock In \emph{2022 {{ACM Conference}} on {{Fairness Accountability}} and
  {{Transparency}}}, pages 789--798, Seoul Republic of Korea. ACM.

\bibitem[{Hofer and Held(2022)}]{HoferHeld2022}
Lisa~J. Hofer and Leonhard Held. 2022.
\newblock \href {https://doi.org/10.48550/arXiv.2207.03199} {Comparing
  {{Confidence Intervals}} for a {{Binomial Proportion}} with the {{Interval
  Score}}}.
\newblock \emph{Preprint}, arXiv:2207.03199.

\bibitem[{Jones et~al.(2019)Jones, Kalbfeld, Hancock, and
  Clark}]{JonesEtAl2019}
Taylor Jones, Jessica~Rose Kalbfeld, Ryan Hancock, and Robin Clark. 2019.
\newblock \href {https://doi.org/10.1353/lan.2019.0042} {Testifying {{While
  Black}}: {{An Experimental Study}} of {{Court Reporter Accuracy}} in
  {{Transcription}} of {{African American English}}}.
\newblock \emph{Language}, 95(2):e216--e252.

\bibitem[{J{\o}rgensen et~al.(2016)J{\o}rgensen, Hovy, and
  S{\o}gaard}]{JorgensenEtAl2016}
Anna J{\o}rgensen, Dirk Hovy, and Anders S{\o}gaard. 2016.
\newblock \href {https://doi.org/10.18653/v1/N16-1130} {Learning a {{POS}}
  tagger for {{AAVE}}-like language}.
\newblock In \emph{Proceedings of the 2016 {{Conference}} of the {{North
  American Chapter}} of the {{Association}} for {{Computational Linguistics}}:
  {{Human Language Technologies}}}, pages 1115--1120, San Diego, California.
  Association for Computational Linguistics.

\bibitem[{Kendall and Farrington(2024)}]{KendallFarrington2024}
Tyler Kendall and Charlie Farrington. 2024.
\newblock {{CORAAL}}.
\newblock https://oraal.github.io/coraal.

\bibitem[{King(2020)}]{King2020}
Sharese King. 2020.
\newblock \href {https://doi.org/10.1146/annurev-linguistics-011619-030556}
  {From {{African American Vernacular English}} to {{African American
  Language}}: {{Rethinking}} the {{Study}} of {{Race}} and {{Language}} in
  {{African Americans}}' {{Speech}}}.
\newblock \emph{Annual Review of Linguistics}, 6(1):285--300.

\bibitem[{Labov(1972)}]{Labov1972}
William Labov. 1972.
\newblock \emph{Language in the {{Inner City}}: {{Studies}} in the {{Black
  English Vernacular}}}.
\newblock Conduct and {{Communication}}. University of Pennsylvania Press,
  Philadelphia.

\bibitem[{Lin et~al.(2025)Lin, Mao, Malfa, Hofmann, de~Wynter, Wang, Chen,
  Wooldridge, Pierrehumbert, and Wei}]{LinEtAl2025}
Fangru Lin, Shaoguang Mao, Emanuele~La Malfa, Valentin Hofmann, Adrian
  de~Wynter, Xun Wang, Si-Qing Chen, Michael Wooldridge, Janet~B.
  Pierrehumbert, and Furu Wei. 2025.
\newblock \href {https://doi.org/10.48550/arXiv.2410.11005} {Assessing
  {{Dialect Fairness}} and {{Robustness}} of {{Large Language Models}} in
  {{Reasoning Tasks}}}.
\newblock \emph{Preprint}, arXiv:2410.11005.

\bibitem[{Martin and Tang(2020)}]{MartinTang2020}
Joshua~L. Martin and Kevin Tang. 2020.
\newblock \href {https://doi.org/10.21437/Interspeech.2020-2893} {Understanding
  {{Racial Disparities}} in {{Automatic Speech Recognition}}: {{The Case}} of
  {{Habitual}} ``be''}.
\newblock In \emph{Interspeech 2020}, pages 626--630. ISCA.

\bibitem[{Masis et~al.(2022)Masis, Neal, Green, and O'Connor}]{MasisEtAl2022}
Tessa Masis, Anissa Neal, Lisa Green, and Brendan O'Connor. 2022.
\newblock Corpus-{{Guided Contrast Sets}} for {{Morphosyntactic Feature
  Detection}} in {{Low-Resource English Varieties}}.
\newblock In \emph{Proceedings of the {{Fifth Workshop}} on {{Natural Language
  Processing}} and {{Computational Social Science}} ({{NLP}}+{{CSS}}) at
  {{EMNLP}} 2022}.

\bibitem[{Miller(2024)}]{Miller2024}
Evan Miller. 2024.
\newblock \href {https://doi.org/10.48550/arXiv.2411.00640} {Adding {{Error
  Bars}} to {{Evals}}: {{A Statistical Approach}} to {{Language Model
  Evaluations}}}.
\newblock \emph{Preprint}, arXiv:2411.00640.

\bibitem[{Mire et~al.(2025)Mire, Aysola, Chechelnitsky, Deas, Zerva, and
  Sap}]{MireEtAl2025}
Joel Mire, Zubin~Trivadi Aysola, Daniel Chechelnitsky, Nicholas Deas, Chrysoula
  Zerva, and Maarten Sap. 2025.
\newblock \href {https://doi.org/10.48550/arXiv.2502.12858} {Rejected
  {{Dialects}}: {{Biases Against African American Language}} in {{Reward
  Models}}}.
\newblock \emph{Preprint}, arXiv:2502.12858.

\bibitem[{{OpenAI}(2025)}]{OpenAI2025GPT41}
{OpenAI}. 2025.
\newblock Introducing {{GPT-4}}.1 in the {{API}}.

\bibitem[{Previlon et~al.(2024)Previlon, Rozet, Gowda, Dyer, Tang, and
  Moeller}]{PrevilonEtAl2024}
Wilermine Previlon, Alice Rozet, Jotsna Gowda, William Dyer, Kevin Tang, and
  Sarah Moeller. 2024.
\newblock \href {https://doi.org/10.31234/osf.io/ph7q8} {Leveraging syntactic
  dependencies in disambiguation: The case of {{African American English}}}.

\bibitem[{Rickford and King(2016)}]{RickfordKing2016}
John~R. Rickford and Sharese King. 2016.
\newblock \href {https://doi.org/10.1353/lan.2016.0078} {Language and
  {{Linguistics}} on {{Trial}}: {{Hearing Rachel Jeantel}} (and {{Other
  Vernacular Speakers}}) in the {{Courtroom}} and {{Beyond}}}.
\newblock \emph{Language}, 92(4):948--988.

\bibitem[{Wu et~al.(2026)Wu, Emami, Hassan, Igbinoba, Idusuyi, Igbinoba,
  Khattak, and {Seyyed-Kalantari}}]{WuEtAl2026}
Huan Wu, Ali Emami, Muhammad~Furquan Hassan, Osaretin Igbinoba, Osakpolor
  Idusuyi, Osamede Igbinoba, Faiza~Khan Khattak, and Laleh {Seyyed-Kalantari}.
  2026.
\newblock \href {https://doi.org/10.48550/arXiv.2607.06845} {{{LLMs Silently
  Correct African American English}}: {{Auditing}} and {{Mitigating Dialect
  Bias}} via {{Activation Steering}}}.
\newblock \emph{Preprint}, arXiv:2607.06845.

\bibitem[{Zhu et~al.(2018)Zhu, Lu, Zheng, Guo, Zhang, Wang, and
  Yu}]{ZhuEtAl2018}
Yaoming Zhu, Sidi Lu, Lei Zheng, Jiaxian Guo, Weinan Zhang, Jun Wang, and Yong
  Yu. 2018.
\newblock \href {https://doi.org/10.1145/3209978.3210080} {Texygen: {{A
  Benchmarking Platform}} for {{Text Generation Models}}}.
\newblock In \emph{The 41st {{International ACM SIGIR Conference}} on
  {{Research}} \& {{Development}} in {{Information Retrieval}}}, pages
  1097--1100, Ann Arbor MI USA. ACM.

\bibitem[{Ziems et~al.(2022)Ziems, Chen, Harris, Anderson, and
  Yang}]{ZiemsEtAl2022}
Caleb Ziems, Jiaao Chen, Camille Harris, Jessica Anderson, and Diyi Yang. 2022.
\newblock \href {https://doi.org/10.48550/arXiv.2204.03031} {{{VALUE}}:
  {{Understanding Dialect Disparity}} in {{NLU}}}.
\newblock \emph{Preprint}, arXiv:2204.03031.

\end{thebibliography}

\appendix

\section{Data preprocessing}
\label{app:preprocessing}

This appendix describes how the CORAAL transcripts were preprocessed before sampling (Section~\ref{sec:data-sampling}).
Following \citet{MasisEtAl2022}, we remove transcript annotations that are not meaningful as language-model input: parenthetical para- and non-linguistic sounds (e.g.\ laughter or coughing), angle-bracketed annotations, unintelligible/inaudible markers, and brackets or quotation marks left over from the transcription conventions.
Speaker identifiers are anonymized (Speaker 1, Speaker 2), and consecutive utterances by the same speaker, which CORAAL often splits across lines (e.g.\ at short prosodic pauses), are merged into a single turn.
Each interview is thus represented as an alternating sequence of speaker turns, one per line.

\section{Classification prompts}
\label{app:classification-prompts}

This appendix contains the prompts used for the feature classification task (Section~\ref{sec:t1-prompting}), along with details on how models were queried.
Under the baseline strategy, the system prompt in Appendix~\ref{app:baseline-prompts} is filled with the feature label, definition and example in Appendix~\ref{app:feature-definitions}.
Under the self-generated strategy, each model condition writes its own system prompt from the meta prompt in Appendix~\ref{app:selfgen-prompts}, using the same feature descriptions.

Each condition is queried once per feature rather than once for all five features: in pilot tests, judging all features in a single call led models to confuse features and spans with one another.
With five votes per feature and segment, each prompting strategy requires 7,500 classification calls (100 segments $\times$ 5 features $\times$ 5 votes $\times$ 3 conditions), plus 15 calls (5 features $\times$ 3 conditions) to generate the self-authored prompts.

\subsection{Baseline system prompt}
\label{app:baseline-prompts}

\begin{promptbox}
You are a sociolinguist annotating African American English (AAE) features in transcribed conversation. You will be shown a short excerpt of five conversational turns from an interview. Decide whether at least one clear instance of the AAE morphosyntactic feature defined below occurs anywhere in the excerpt.

Feature: <feature label>
Definition: <feature definition>
Example: <feature example>

Annotation guidelines:
- Judge the feature strictly by the definition given above.
- Use the full conversational context to annotate the excerpt.
- The feature is present if at least one clear instance occurs in any turn of the excerpt.
- If the feature is present, quote the exact minimal substring from the excerpt that contains it. For an absence-based feature, quote the phrase containing the gap. If the feature is not present, use an empty string for the span.

Return only a JSON object with a boolean 'present' field and a 'span' string. Do not include any other text.
\end{promptbox}

\subsection{Feature definitions}
\label{app:feature-definitions}

The label, definition and example inserted into the \texttt{<feature label>}, \texttt{<feature definition>} and \texttt{<feature example>} slots of both prompt templates are listed below for each feature.

\begin{promptbox}
Feature: habitual be / aspectual be / invariant be
Definition: Habitual be occurs when an uninflected form of the verb 'to be' is used to mark a habitual or repeated action.
Example: She be telling people she eight.

Feature: copula deletion / copula absence / null copula / zero copula
Definition: Copula deletion occurs when a form of the verb 'to be' (i.e. the copula) is omitted.
Example: He sleeping in that car.

Feature: third-person singular -s deletion / absence
Definition: Third person singular -s deletion refers to the deletion of the -s suffix from a present-tense verb with a third-person singular subject.
Example: When he come down here.

Feature: is/was generalization
Definition: Is/was generalization refers to the use of the singular verb forms 'is' or 'was' with a plural subject.
Example: What they was doing?

Feature: multiple negation / negative concord
Definition: Multiple negation refers to the use of two or more negative markers within a single clause expressing a single negation, rather than canceling each other out.
Example: Can't nobody tell you it wasn't meant for you.
\end{promptbox}

\subsection{Self-generated prompts: meta prompt}
\label{app:selfgen-prompts}

\begin{promptbox}
You are an expert prompt engineer and sociolinguist. Your task is to write the best possible system prompt for a language model that will annotate a single African American English (AAE) morphosyntactic feature in transcribed conversation.

The annotation model that receives your prompt will be shown a short excerpt of five conversational turns from an interview. It must decide whether at least one clear instance of the target feature occurs anywhere in the excerpt, and return a JSON object with a boolean 'present' field and a 'span' string quoting the exact triggering text (an empty string when the feature is absent).

The target feature for this prompt is:
Feature: <feature label>
Definition: <feature definition>
Example: <feature example>

Write a system prompt that makes the annotation model as accurate as possible on this feature. Precision is the priority: on this task the model currently over-predicts the feature and produces many false positives, so your prompt should help it avoid flagging cases that merely resemble the feature but do not satisfy the definition. The model should mark the feature present only when a specific, identifiable instance is clearly supported by the text, without overlooking real instances.

Output only the system prompt text itself, with no preamble, commentary, or surrounding quotation marks.
\end{promptbox}

\section{Supervised baseline}
\label{app:supervised-baseline}

This appendix details how the \citet{MasisEtAl2022} classifier was adapted to the five features and to segment-level prediction (Section~\ref{sec:t1-supervised-baseline}).
We use seven of the 17 binary heads of the \citet{MasisEtAl2022} classifier.
Multiple negation maps onto three heads (multiple negation, inverted negative auxiliary and non-inverted negative auxiliary), which we pool (cf.\ Section~\ref{sec:features}).
Because the classifier scores individual sentences of at most 64 tokens, we split each turn into sentences and take the maximum score across sentences (and, for multiple negation, across its three heads), and then across the five turns of a segment.

\section{Pairwise contrast tests}
\label{app:pairwise-tests}

This appendix reports the full results of the five pre-specified macro F1 contrasts summarized in Figure~\ref{fig:t1-f1-contrasts} (Section~\ref{sec:t1-evaluation}) in Table~\ref{tab:app-pairwise-tests}.
For each contrast, we compute the F1 difference on each of the 10,000 paired bootstrap resamples.
Intervals are percentile intervals at the family-wise $1 - \alpha/m = 99\%$ level ($m = 5$), and p-values are two-sided bootstrap p-values, Bonferroni-adjusted by multiplying by $m$.
An interval excluding zero thus marks a significant contrast.

\section{Per-feature performance}
\label{app:per-feature-table}

\begin{figure*}[!tp]
\centering
\includegraphics[width=\textwidth]{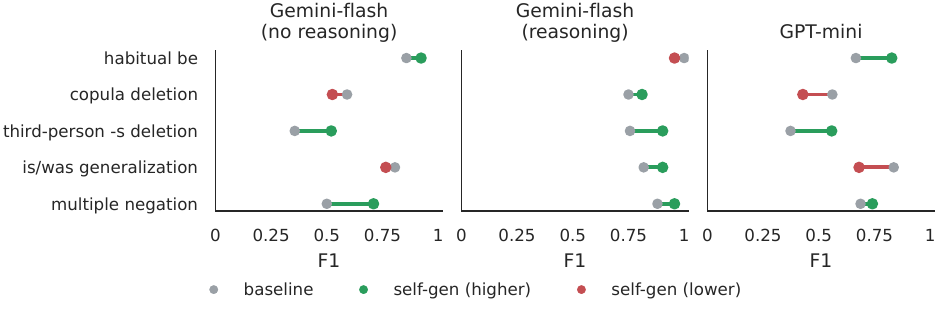}
\caption{Per-feature F1 shifts from the baseline to the self-generated prompt for the three LLM conditions.}
\label{fig:t1-prompt-comparison}
\end{figure*}

This appendix reports per-feature precision, recall and F1 for every condition and prompting strategy, with 95\% BCa bootstrap intervals (Tables~\ref{tab:app-per-feature-habitual-be}--\ref{tab:app-per-feature-multiple-negation}), as well as the per-feature F1 change from the baseline to the self-generated prompt (Figure~\ref{fig:t1-prompt-comparison}).

The macro-level prompt effects reported in Section~\ref{sec:t1-overall} are not uniform across features.
For GPT-mini, the near-zero macro change appears to be a redistribution, since the precision-focused prompt raises F1 where the model had been overpredicting (third-person -s deletion +0.18, habitual be +0.16), but lowers it on is/was generalization (-0.16) and copula deletion (-0.13).
The two Gemini conditions, on the other hand, shift almost entirely upward, improving most noticeably for multiple negation and third-person -s deletion, which are the same features for which Figure~\ref{fig:t1-per-feature-f1} indicated the largest reasoning advantage.

\section{Span validation}
\label{sec:t1-span-validation}

This appendix analyzes the text spans quoted in the models' positive predictions, to assess whether false positives reflect misjudgments or hallucinated evidence (Section~\ref{sec:t1-per-feature}).
We match each span quoted in a positive prediction against the turn-level gold annotations and assign it to one of four categories: (i) in a turn annotated for that feature, (ii) in another turn, (iii) in the segment but across turns, or (iv) not found in the segment.
Because the gold labels are turn-level, this checks localization at the level of the turn rather than the phrase.

For false positives, which by construction have no annotated turn, we distinguish misjudgment (a real span labeled incorrectly) from hallucination (a quoted span not present in the transcript).
Hallucination was rare: across all conditions, only about 4\% of false positive spans (77 in total) failed to match the transcript.
On closer inspection, these were largely non-verbatim renderings of real speech rather than invented text, often assembled from several turns (e.g.\ a concatenation of \lingex{she feel\ldots\ she don't\ldots\ she come}).
False positives were consequently almost entirely misjudgments, for instance tagging spans like \lingex{we was} (is/was generalization) or \lingex{it was four of us} (a cleft construction) as copula deletion.

For true positives, Table~\ref{tab:t1-span-validation} shows the proportion of quoted spans that fall in the gold-annotated turn.
This rate is near-perfect for features carrying a noticeable lexical element, such as multiple negation (0.98--1.00), is/was generalization (0.87--0.98) and habitual be (0.91--1.00), but lower for third-person -s deletion (0.83--0.93) and copula deletion (0.59--0.92).
These two features are characterized by the absence of a particular linguistic element, which might make them harder to pinpoint in a stretch of speech; they also showed lower precision (Section~\ref{sec:t1-per-feature}).

\section{Next-turn generation prompt}
\label{app:t2-prompt}

This appendix contains the system prompt used for the next-turn generation task (Section~\ref{sec:t2-generation-setup}).

\begin{promptbox}
You are continuing a transcript of a naturally occurring conversational dialogue from a recorded interview. Two speakers are talking with each other and take turns in the exchange. Given the preceding turns of the conversation, produce the single next turn that the indicated speaker would most naturally say at this point in the dialogue.

You will also be given a brief description of what the next turn should convey. Use it to decide the content of the turn, but phrase the turn the way the speaker would naturally say it in this conversation.

Output only the words of that turn: no speaker label, no quotation marks, and no description or commentary.
\end{promptbox}

\section{Pragmatic quality}
\label{app:pragmatic-quality}

This appendix contains additional analyses of the 200 pairwise preference trials (Sections~\ref{sec:t2-evaluation} and~\ref{sec:t2-quality}).

\begin{figure*}[!tp]
\centering
\includegraphics[width=0.5\textwidth]{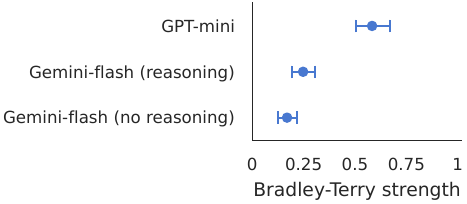}
\caption{Bradley-Terry latent quality strength per condition, with 95\% bootstrap confidence intervals.}
\label{fig:t2-bradley-terry}
\end{figure*}

Figure~\ref{fig:t2-bradley-terry} summarizes the trials through a Bradley-Terry fit \citep{BradleyTerry1952}, which estimates each condition's relative strength as a single latent parameter, counting ties as half a win to each side, with 95\% bootstrap intervals.
The ordering matches the raw winrates (Table~\ref{tab:t2-win-rate}): GPT-mini (0.58 [0.50, 0.67]), Gemini-flash (reasoning) (0.25 [0.19, 0.31]) and Gemini-flash (no reasoning) (0.17 [0.13, 0.22]), with the intervals of the two Gemini-flash conditions partly overlapping.

Table~\ref{tab:t2-sign-tests} gives the pairwise tests: two-sided exact binomial sign tests over each pair's non-tied trials against a null winrate of 0.5, Bonferroni-corrected ($m = 3$), with exact Clopper-Pearson intervals \citep{HoferHeld2022} at the family-wise level.

\clearpage
\makeatletter
\setlength{\@dblfptop}{0pt}
\setlength{\@dblfpsep}{12pt plus 2pt}
\setlength{\@dblfpbot}{0pt plus 1fil}
\makeatother

\begin{table*}[!tp]
\centering
\small
\begin{tabular}{l c c c}
\toprule
Contrast & $\Delta$ F1 & 99\% FW CI & Adjusted $p$ \\
\midrule
Gemini-flash (reasoning) vs.\ (no reasoning), baseline   & 0.219 & [0.147, 0.307] & $<.001$ \\
Gemini-flash (reasoning) vs.\ (no reasoning), self-gen.  & 0.218 & [0.138, 0.310] & $<.001$ \\
Gemini-flash (reasoning), self-gen.\ vs.\ baseline       & 0.065 & [0.011, 0.120] & .016 \\
Gemini-flash (no reasoning), self-gen.\ vs.\ baseline    & 0.066 & [0.006, 0.129] & .026 \\
GPT-mini, self-gen.\ vs.\ baseline                       & 0.022 & [-0.061, 0.101] & 1.000 \\
\bottomrule
\end{tabular}
\caption{The five pre-specified paired contrasts on macro F1, with point estimates, 99\% family-wise percentile bootstrap intervals and Bonferroni-adjusted bootstrap p-values ($m = 5$).}
\label{tab:app-pairwise-tests}
\end{table*}

\begin{table*}[!tp]
\centering
\small
\begin{tabular}{l c c c}
\toprule
Condition & Precision & Recall & F1 \\
\midrule
Gemini-flash (reasoning), baseline & 1.000 [1.000, 1.000]$^c$ & 1.000 [1.000, 1.000]$^c$ & 1.000 [1.000, 1.000]$^c$ \\
Gemini-flash (reasoning), self-gen. & 1.000 [1.000, 1.000]$^c$ & 0.917 [0.518, 1.000] & 0.957 [0.695, 1.000] \\
Gemini-flash (no reasoning), baseline & 0.750 [0.455, 0.923] & 1.000 [1.000, 1.000]$^c$ & 0.857 [0.625, 0.960] \\
Gemini-flash (no reasoning), self-gen. & 0.857 [0.538, 1.000] & 1.000 [1.000, 1.000]$^c$ & 0.923 [0.700, 1.000] \\
GPT-mini, baseline & 0.500 [0.273, 0.688] & 1.000 [1.000, 1.000]$^c$ & 0.667 [0.429, 0.815] \\
GPT-mini, self-gen. & 0.706 [0.429, 0.889] & 1.000 [1.000, 1.000]$^c$ & 0.828 [0.600, 0.941] \\
\citet{MasisEtAl2022} & 0.917 [0.500, 1.000] & 0.917 [0.500, 1.000] & 0.917 [0.667, 1.000] \\
\bottomrule
\end{tabular}
\caption{Per-feature precision, recall and F1 for habitual be ($12$ gold-positive segments), with 95\% confidence intervals ($^c$ marks an interval collapsed at the 0/1 boundary).}
\label{tab:app-per-feature-habitual-be}
\end{table*}

\begin{table*}[!tp]
\centering
\small
\begin{tabular}{l c c c}
\toprule
Condition & Precision & Recall & F1 \\
\midrule
Gemini-flash (reasoning), baseline & 0.600 [0.452, 0.727] & 1.000 [1.000, 1.000]$^c$ & 0.750 [0.623, 0.842] \\
Gemini-flash (reasoning), self-gen. & 0.682 [0.525, 0.810] & 1.000 [1.000, 1.000]$^c$ & 0.811 [0.689, 0.895] \\
Gemini-flash (no reasoning), baseline & 0.448 [0.321, 0.577] & 0.867 [0.692, 0.963] & 0.591 [0.460, 0.705] \\
Gemini-flash (no reasoning), self-gen. & 0.420 [0.283, 0.561] & 0.700 [0.515, 0.846] & 0.525 [0.384, 0.652] \\
GPT-mini, baseline & 0.442 [0.308, 0.580] & 0.767 [0.576, 0.895] & 0.561 [0.423, 0.681] \\
GPT-mini, self-gen. & 0.462 [0.261, 0.652] & 0.400 [0.226, 0.585] & 0.429 [0.261, 0.588] \\
\citet{MasisEtAl2022} & 0.724 [0.531, 0.864] & 0.700 [0.517, 0.848] & 0.712 [0.566, 0.830] \\
\bottomrule
\end{tabular}
\caption{Per-feature precision, recall and F1 for copula deletion ($30$ gold-positive segments), with 95\% confidence intervals ($^c$ marks an interval collapsed at the 0/1 boundary).}
\label{tab:app-per-feature-copula-deletion}
\end{table*}

\begin{table*}[!tp]
\centering
\small
\begin{tabular}{l c c c}
\toprule
Condition & Precision & Recall & F1 \\
\midrule
Gemini-flash (reasoning), baseline & 0.609 [0.381, 0.792] & 1.000 [1.000, 1.000]$^c$ & 0.757 [0.552, 0.884] \\
Gemini-flash (reasoning), self-gen. & 0.824 [0.550, 0.950] & 1.000 [1.000, 1.000]$^c$ & 0.903 [0.714, 0.974] \\
Gemini-flash (no reasoning), baseline & 0.220 [0.125, 0.339] & 0.929 [0.571, 1.000] & 0.356 [0.219, 0.500] \\
Gemini-flash (no reasoning), self-gen. & 0.361 [0.211, 0.529] & 0.929 [0.571, 1.000] & 0.520 [0.340, 0.679] \\
GPT-mini, baseline & 0.230 [0.132, 0.347] & 1.000 [1.000, 1.000]$^c$ & 0.373 [0.234, 0.516] \\
GPT-mini, self-gen. & 0.414 [0.238, 0.600] & 0.857 [0.545, 1.000] & 0.558 [0.364, 0.727] \\
\citet{MasisEtAl2022} & 0.750 [0.250, 1.000] & 0.429 [0.167, 0.692] & 0.545 [0.250, 0.769] \\
\bottomrule
\end{tabular}
\caption{Per-feature precision, recall and F1 for third-person singular -s deletion ($14$ gold-positive segments), with 95\% confidence intervals ($^c$ marks an interval collapsed at the 0/1 boundary).}
\label{tab:app-per-feature-third-person-s-deletion}
\end{table*}

\begin{table*}[!tp]
\centering
\small
\begin{tabular}{l c c c}
\toprule
Condition & Precision & Recall & F1 \\
\midrule
Gemini-flash (reasoning), baseline & 0.711 [0.543, 0.838] & 0.964 [0.800, 1.000] & 0.818 [0.690, 0.903] \\
Gemini-flash (reasoning), self-gen. & 0.824 [0.654, 0.929] & 1.000 [1.000, 1.000]$^c$ & 0.903 [0.792, 0.963] \\
Gemini-flash (no reasoning), baseline & 0.692 [0.525, 0.824] & 0.964 [0.790, 1.000] & 0.806 [0.667, 0.893] \\
Gemini-flash (no reasoning), self-gen. & 0.650 [0.486, 0.789] & 0.929 [0.762, 1.000] & 0.765 [0.630, 0.862] \\
GPT-mini, baseline & 0.852 [0.667, 0.960] & 0.821 [0.636, 0.935] & 0.836 [0.698, 0.923] \\
GPT-mini, self-gen. & 0.842 [0.600, 0.956] & 0.571 [0.379, 0.750] & 0.681 [0.500, 0.816] \\
\citet{MasisEtAl2022} & 0.769 [0.567, 0.905] & 0.714 [0.514, 0.857] & 0.741 [0.583, 0.857] \\
\bottomrule
\end{tabular}
\caption{Per-feature precision, recall and F1 for is/was generalization ($28$ gold-positive segments), with 95\% confidence intervals ($^c$ marks an interval collapsed at the 0/1 boundary).}
\label{tab:app-per-feature-is-was-generalization}
\end{table*}

\begin{table*}[!tp]
\centering
\small
\begin{tabular}{l c c c}
\toprule
Condition & Precision & Recall & F1 \\
\midrule
Gemini-flash (reasoning), baseline & 0.786 [0.467, 0.944] & 1.000 [1.000, 1.000]$^c$ & 0.880 [0.636, 0.972] \\
Gemini-flash (reasoning), self-gen. & 0.917 [0.529, 1.000] & 1.000 [1.000, 1.000]$^c$ & 0.957 [0.697, 1.000] \\
Gemini-flash (no reasoning), baseline & 0.333 [0.176, 0.500] & 1.000 [1.000, 1.000]$^c$ & 0.500 [0.300, 0.667] \\
Gemini-flash (no reasoning), self-gen. & 0.550 [0.316, 0.765] & 1.000 [1.000, 1.000]$^c$ & 0.710 [0.480, 0.867] \\
GPT-mini, baseline & 0.524 [0.300, 0.733] & 1.000 [1.000, 1.000]$^c$ & 0.688 [0.462, 0.848] \\
GPT-mini, self-gen. & 0.625 [0.333, 0.846] & 0.909 [0.460, 1.000] & 0.741 [0.500, 0.900] \\
\citet{MasisEtAl2022} & 0.727 [0.349, 0.933] & 0.727 [0.357, 0.933] & 0.727 [0.444, 0.900] \\
\bottomrule
\end{tabular}
\caption{Per-feature precision, recall and F1 for multiple negation ($11$ gold-positive segments), with 95\% confidence intervals ($^c$ marks an interval collapsed at the 0/1 boundary).}
\label{tab:app-per-feature-multiple-negation}
\end{table*}

\begin{table*}[!tp]
\centering
\small
\begin{tabular}{l c c c c c c}
\toprule
Condition & hab.\ be & cop.\ del. & 3sg -s del. & is/was gen. & mult.\ neg. & macro avg. \\
\midrule
Gemini-flash (reasoning), baseline & 0.98 & 0.88 & 0.93 & 0.98 & 1.00 & 0.95 \\
Gemini-flash (reasoning), self-gen. & 1.00 & 0.92 & 0.86 & 0.96 & 1.00 & 0.95 \\
Gemini-flash (no reasoning), baseline & 0.93 & 0.63 & 0.83 & 0.98 & 0.98 & 0.87 \\
Gemini-flash (no reasoning), self-gen. & 0.92 & 0.76 & 0.84 & 0.94 & 1.00 & 0.89 \\
GPT-mini, baseline & 0.92 & 0.79 & 0.91 & 0.96 & 1.00 & 0.92 \\
GPT-mini, self-gen. & 0.91 & 0.59 & 0.83 & 0.87 & 1.00 & 0.84 \\
\bottomrule
\end{tabular}
\caption{Proportion of true-positive spans correctly located in a gold-annotated turn, per condition and feature, with a macro average across features.}
\label{tab:t1-span-validation}
\end{table*}

\begin{table*}[!tp]
\centering
\small
\setlength{\tabcolsep}{4pt}
\begin{tabular}{l c c c c c c}
\toprule
Comparison (A vs.\ B) & A wins & B wins & Ties & Winrate A & Raw $p$ & Adj.\ $p$ \\
\midrule
Gemini-flash (no reasoning) vs.\ Gemini-flash (reasoning) & 12 & 27 & 31 & 0.31 [0.15, 0.51] & .024 & .071 \\
Gemini-flash (no reasoning) vs.\ GPT-mini & 8 & 38 & 12 & 0.17 [0.06, 0.35] & $<.001$ & $<.001$ \\
Gemini-flash (reasoning) vs.\ GPT-mini & 11 & 42 & 19 & 0.21 [0.09, 0.37] & $<.001$ & $<.001$ \\
\bottomrule
\end{tabular}
\caption{Pairwise exact binomial sign tests over each pair's non-tied trials, treated as one Bonferroni family ($m = 3$). The Clopper-Pearson interval on the win rate of condition A is reported at the family-wise ($1 - \alpha/m \approx 98.3\%$) level.}
\label{tab:t2-sign-tests}
\end{table*}

\end{document}